\documentclass[a4paper,twoside]{article}

\usepackage{epsfig}
\usepackage{subfigure}
\usepackage{calc}
\usepackage{amssymb}
\usepackage{amstext}
\usepackage{amsmath}
\usepackage{amsthm}
\usepackage{multicol}
\usepackage{pslatex}
\usepackage{apalike}

\usepackage{cite}
\usepackage{graphicx}
\usepackage{csvsimple}
\usepackage[export]{adjustbox}
\usepackage{multirow}
\usepackage[table,xcdraw]{xcolor}
\usepackage{float}

\usepackage{SCITEPRESS}     

\begin{document}

\title{Evaluating Deep Convolutional Neural Networks for Material Classification}

\author{\authorname{Grigorios Kalliatakis\sup{1}, Georgios Stamatiadis\sup{1}, Shoaib Ehsan\sup{1}, Ales Leonardis\sup{2}, Juergen Gall\sup{3}, Anca Sticlaru\sup{1} and Klaus D. McDonald-Maier\sup{1}}
\affiliation{\sup{1}School of Computer Science and Electronic Engineering, University of Essex, Colchester, UK}
\affiliation{\sup{2} School of Computer Science, University of Birmingham, Birmingham, UK}
\affiliation{\sup{3} Institute of Computer Science, University of Bonn, Bonn, Germany}
\email{\{gkallia, gstama, sehsan, asticl, kdm\}@essex.ac.uk, a.leonardis@cs.bham.ac.uk, gall@iai.uni-bonn.de}
}


\abstract{Determining the material category of a surface from an image is a demanding task in perception that is drawing increasing attention. Following the recent remarkable results achieved for image classification and object detection utilising Convolutional Neural Networks (CNNs), we empirically study material classification of everyday objects employing these techniques. More specifically, we conduct a rigorous evaluation of how state-of-the art CNN architectures compare on a common ground over widely used material databases. Experimental results on three challenging material databases show that the best performing CNN architectures can achieve up to 94.99\% mean average precision when classifying materials.}

\onecolumn \maketitle \normalsize \vfill

\section{\uppercase{Introduction}}
\label{sec:introduction}

\noindent Image classification and object detection have been active areas of research during the last few years \cite{10,27,28}. Initially, handcrafted approaches, such as Bag-of-Visual-Words (BoVW) \cite{1}, were employed that yielded reasonably good results for these two tasks. However, the emergence of Convolutional Neural Networks (CNNs) \cite{2} for solving these vision based problems has changed the scenario altogether by comprehensively outperforming the handcrafted approaches \cite{7,9,5}. While the morphology of these networks remains handcrafted, the accommodation of a large number of parameters trained from data and numerous layers of non-linear feature extractors have lead the researchers to term them as deep representations. After setting the performance benchmark for image classification and object detection tasks \cite{3,4}, these deep architectures are now finding their way into a number of vision based applications \cite{7,29,8,9,5,6}.

\begin{figure*}[!t]
\centering
\includegraphics[height=6.5cm,width=13.25cm,keepaspectratio,center]{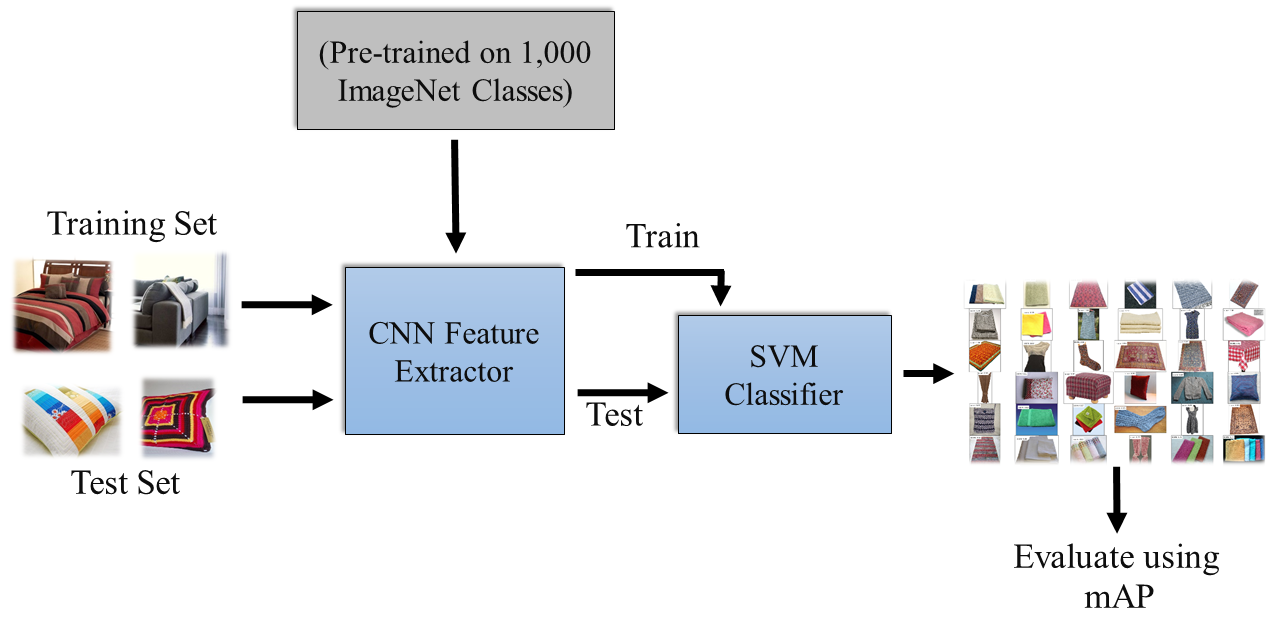}
\caption{An overview of the material classification pipeline used for our experiments. The training and test datasets, along with the learning procedure, are fixed as different CNN architectures (pre-trained on 1000 ImageNet classes) are plugged into the pipeline, one at a time, to test their performance using mean average precision (mAP).}
\label{overview}
\end{figure*}

One such application is classification of materials from their appearance utilising a single image. Indeed, perception of recognized surface material plays a major role in scene understanding and has a wide range of applications, including robotics. Material classification in the wild is considered a challenging problem due to the fact that materials regularly exhibit large intra-class and inter-class variability. This particular topic has received attention fairly recently and a handful of computer vision systems have been explicitly constructed to recognize materials in the wild so far. In the early work on image texture analysis by \cite{11}, the CUReT dataset was introduced which was generated in a restricted environment and only encompassed flat texture patches. Over 95\% classification accuracy was reported on the CUReT dataset by \cite{12}. On the contrary, only 23\% accuracy was accomplished on the more challenging Flickr material dataset (FMD) \cite{13}. The work of \cite{14}, in which they presented a number of new features for classifying materials, achieved 45\% accuracy on FMD. This was improved by \cite{15} that achieved 54\% classification accuracy by expanding more on features. Finally, \cite{16} developed a CNN and improved Fisher vector (IFV) classifier that achieved state-of-the-art results on FMD and KTH-TIPS2 \cite{17}.

It is, however, still unclear how different CNN methods compare with each other for the material classification task as previous studies did not test these deep architectures extensively on common ground \cite{24}. Since the performance of computer vision systems depends significantly on implementation details \cite{20}, it is important to take into account factors, such as the use of very large scale datasets, GPU computation and data augmentation, when comparing CNN methods for material classification. To our knowledge, this particular work is the first attempt to do a systematic and rigorous evaluation of the state-of-the-art CNN architectures for material classification in the wild. Considering the effect of different design and implementation choices allowed a fair, unbiased comparison on a common ground-something that has been largely missing so far in the literature.

More specifically, we perform thorough assessment of the state-of-the-art CNN architectures utilising three widely used materials databases (FMD \cite{13}, MINC-2500 \cite{24} and ImageNet materials \cite{15}) while identifying and disclosing important implementation details. For performing the large set of experiments, we partly followed the approach of \cite{20} which was used for comparing CNN architectures for recognition of object categories. We, on the other hand, tackle material classification in this particular work, an entirely different problem from \cite{20}. Our experimental results on three challenging materials databases show that the best performing CNN architectures can achieve up to 94.99\% mean average precision when classifying materials.

The rest of the paper is organised as follows. Section 2 gives details of the material classification pipeline used for our experiments. The evaluation results for state-of-the-art CNN architectures employing three widely used materials databases of real-world images are presented in Section 3. Section 4 performs a cross-dataset analysis for real-world images. Finally, conclusions are given in Section 5.

\begin{table*}[t]
\centering
\caption{An overview of the three material databases used for experiments.}
\label{my-label}
\begin{tabular}{c|c|c|c|}
\cline{2-4}
                                           & \textbf{FMD} & \textbf{ImageNet7} & \textbf{MINC-2500} \\ \hline
\multicolumn{1}{|c|}{Categories}           & 10           & 7                  & 23                 \\ \hline
\multicolumn{1}{|c|}{Samples per category} & 100          & 100                & 2500               \\ \hline
\multicolumn{1}{|c|}{Material Samples}     & 1000         & 1000               & 2500               \\ \hline
\multicolumn{1}{|c|}{Total image number}   & 1000         & 7000               & 57500              \\ \hline
\end{tabular}
\end{table*}

\section{\uppercase{MATERIAL CLASSIFICATION PIPELINE}}
\noindent An illustration of the material classification pipeline used for our experiments is given in Figure 1. In this pipeline, every block is fixed except the feature extractor as different CNN architectures (pre-trained on 1000 ImageNet classes) are plugged in, one at a time, to compare their performance utilising mean average precision (mAP). Given a training dataset Tr consisting of m material categories, a test dataset Ts comprising unseen images of the material categories given in Tr, and a set of n pre-trained CNN architectures (C1,...Cn), the pipeline operates as follows: The training dataset Tr is used as input to the first CNN architecture C1. The output of C1 is then utilised to train m SVM classifiers. Once trained, the test dataset Ts is employed to assess the performance of the material classification pipeline using mAP. The training and testing procedures are then repeated after replacing C1 with the second CNN architecture C2 to evaluate the performance of
the material classification pipeline. For a set of n pre-trained CNN architectures, the training and testing processes are repeated n times. Since the whole pipeline is fixed (including the training and test datasets, learning procedure and evaluation protocol) for all n CNN architectures, the differences in the performance of the material classification pipeline can be attributed to the specific CNN architectures used.

The CNN-F architecture is similar to the one used by \cite{3}. On the other hand, the CNN-M architecture is similar to the one employed by \cite{4}, whereas the CNN-S architecture is related to the 'accurate' network from the OverFeat package \cite{5}. All these baseline CNN architectures are built on the Caffe framework \cite{22} and are pre-trained on ImageNet \cite{23}. Each network comprises 5 convolutional and 3 fully connected layers for a total of 8 learnable layers. For further design and implementation details for these architectures, please see Table 1 in \cite{20}. Please note that the results of the penultimate layer (layer 7) are used for the SVM classifier in this particular work. Each test yields a feature vector of 4096 dimensions per image. The CNN-M is also tested in situations when the feature dimensionality is reduced to 2048, 1024, and 128, and in cases where the images are turned into grey scales.

Three different types of data augmentation are used: 1) No augmentation, where a 224 x 224 crop is taken from the image (image is downsized to224 pixels in the smallest dimension); 2) Flip augmentation, where the image is mirrored along the y-axis; and 3) Crop and Flip augmentation, where the four corners of the image and the center and their flips are taken and rescaled down to 256 pixels on the smallest side. In terms of collation, there are four types used: 1) No collation, where the additional crops generated by the various augmentation methods are returned as extra features; 2) Sum pooling is used over the generated crops for each image; 3) Max, where max pooling is used; and 4) Stack, where the crops generated are stacked and thus yield feature vectors of more dimensions per image.

\section{\uppercase{PERFORMANCE COMPARISON OF CNN ARCHITECTURES}}
\noindent This section presents the results for the three baseline CNN architectures, with different data augmentation strategies, for the material classification task when trained and tested on real- world images. The evaluation procedure is divided into three different sets of experiments, each one employing a different, widely used materials database consisting of real-world images related to specific material categories. In each case, the employed materials database is used for generating the training and testing datasets which implies no cross-dataset analysis for these particular sets of experiments. This approach is used to obtain comparison results across all available material categories for each benchmark database, thus complementing the previous studies in the literature on these databases.

\begin{table*}[t]
\centering
\caption{Material classification results with real-world images. Both training and testing are performed using the same database. Bold font highlights the leading mean result for every database. Three data augmentation strategies are used for both training and testing: 1) no augmentation (denoted Image Aug=-), 2) flip augmentation (denoted Image Aug=(F)), 3) crop and flip (denoted Image Aug=(C)). Augmented images are used as stand-alone samples (f), or by combining the corresponding descriptors using sum (s) or max (m) pooling or stacking (t). Here, GS denotes gray scale. The same symbols for data augmentation options and gray scale are used in the rest of the paper.}
\label{my-label}
\begin{tabular}{cccc|ccc}
\hline
\rowcolor[HTML]{C0C0C0} 
\cellcolor[HTML]{C0C0C0}                                  & \multicolumn{3}{c|}{\cellcolor[HTML]{C0C0C0}}                                      & \textbf{MINC-2500} & \textbf{ImageNet7} & \textbf{FMD}   \\ \cline{5-7} 
\rowcolor[HTML]{C0C0C0} 
\multirow{-2}{*}{\cellcolor[HTML]{C0C0C0}\textbf{Method}} & \multicolumn{3}{c|}{\multirow{-2}{*}{\cellcolor[HTML]{C0C0C0}\textbf{Image Aug.}}} & \textbf{mAP}       & \textbf{mAP}       & \textbf{mAP}   \\
(a) CNN F                                                 & (C)                          & f                        & s                        & 91.68              & 67.68              & 59.39          \\
(b) CNN S                                                 & (C)                          & f                        & s                        & 92.98              & 70.47              & 64.44          \\ \hline
(c) CNN M                                                 & -                            &                          &                          & 92.14              & 70.67              & 60.72          \\
(d) CNN M                                                 & (C)                          & f                        & s                        & 92.64              & 72.50              & 62.72          \\
(e) CNN M                                                 & (C)                          & f                        & m                        & 92.85              & 73.28              & 62.97          \\
(f) CNN M                                                 & (C)                          & s                        & s                        & 93.17              & 71.86              & 62.57          \\
(g) CNN M                                                 & (C)                          & t                        & t                        & \textbf{94.99}     & 73.73              & \textbf{64.40} \\
(h) CNN M                                                 & (C)                          & f                        & -                        & 91.23              & 69.87              & 58.88          \\
(i) CNN M                                                 & (F)                          & f                        & -                        & 91.94              & 71.08              & 60.37          \\
(j) CNN M GS                                              & -                            &                          &                          & 90.54              & 67.31              & 52.38          \\
(k) CNN M GS                                              & (C)                          & f                        & s                        & 90.87              & 67.48              & 59.23          \\ \hline
(l) CNN M 2048                                            & (C)                          & f                        & s                        & 93.34              & 72.55              & 62.45          \\
(m) CNN M 1024                                            & (C)                          & f                        & s                        & 93.61              & 73.09              & 61.92          \\
(n) CNN M 128                                             & (C)                          & f                        & s                        & 92.74              & \textbf{74.97}     & 48.18          \\ \hline
\end{tabular}
\end{table*}

\subsection{Material Databases}
\noindent Three different databases are used in our experiments: 1) Flickr Material Database (FMD) \cite{13}, 2) ImageNet7 dataset \cite{15} which was derived from ImageNet \cite{23} by collecting 7 common material categories, and 3) MINC-2500 which is a patch classification dataset with 2500 samples per category \cite{24}. Table 1 gives an overview of the three different material databases used for these experiments. As evident, all three databases consist of neither the same number of images nor categories between them. For this specific reason and in order to keep the tests on a common base, we consider the first half of the images enclosed in each database category as positive training samples and the other half for testing. Regarding negative training samples, the first 10\% of the total images per category are aggregated in order to generate the negative training subset. Finally, a dataset \cite{21} containing 1414 random images is utilised and kept constant as the negative test data of our system for all the experiments that follow. In total, 14 different variants of the baseline CNN architectures with different data augmentation strategies are compared on FMD, ImageNet7 and MINC-2500.

\begin{figure*}[t]
\centering
\includegraphics[height=9.5cm,width=15.25cm,keepaspectratio,center]{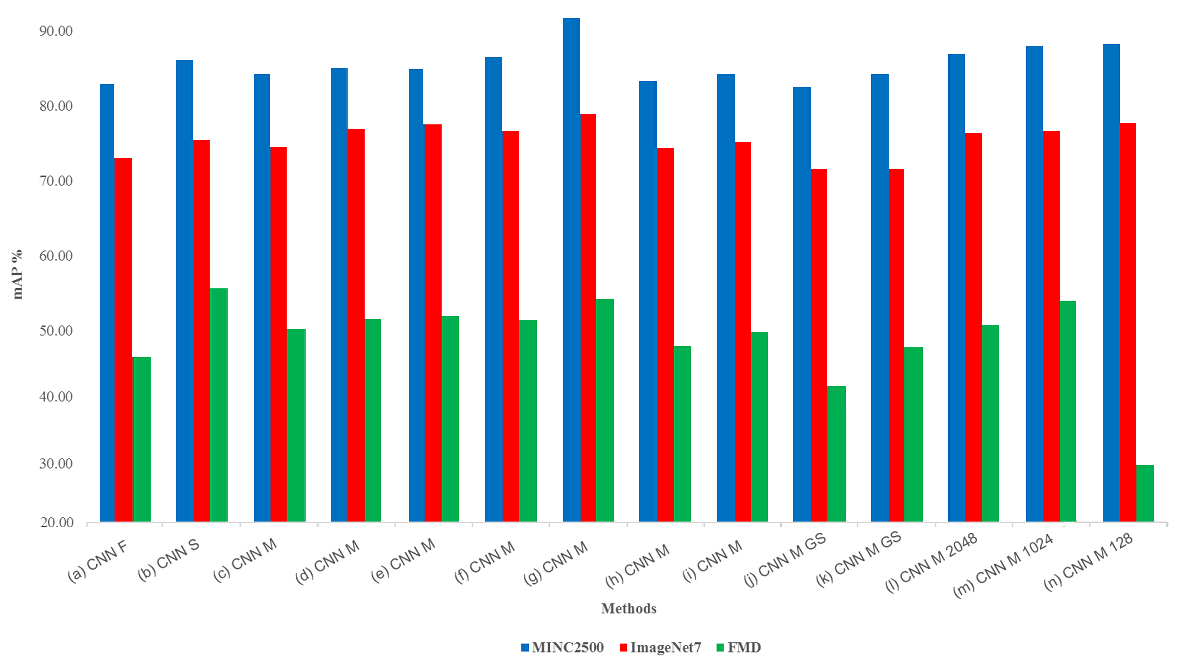}
\caption{Comparison of CNN architectures in terms of mAP for six overlapping categories (fabric, glass, metal, paper, plastic and wood) of FMD, MINC-2500, ImageNet7 databases.}
\label{mAP}
\end{figure*}

\subsection{Results and Discussion}
\noindent Table 2 shows the results for the three databases for state-of-the-art CNN architectures with different data augmentation strategies. It is evident that the Medium CNN architecture with crop and flip augmentation enabled, whereas the corresponding descriptors using stacking both in training and testing, used for the samples of augmented images, performs the best for both MINC-2500 and FMD. On the other hand, the Medium CNN architecture, including lower dimensional full 7 layers of 128 dimensions, with crop and flip augmentation enabled, when the corresponding descriptors using sum-pooling are utilised for the samples of augmented images in testing only, performs the best in the case of ImageNet7. Compared to \cite{8}, where 85.0\% mean accuracy was achieved on MINC (considering only the common categories with FMD) using the AlexNet \cite{3}, here an average of 92.48\% on MINC-2500 is achieved across all considered CNN architectures. Regarding the FMD database, \cite{14} with their optimal feature set managed 44.6\% recognition rate, while in \cite{29} 54\% accuracy is achieved with their extended kernel descriptors. In our case, an improved mAP of 60\% (on average) is achieved for the 14 different CNN configurations that we tested. Finally, the creators of the ImageNet7 \cite{15} reported 60\% recognition accuracy on their database by combining all 5 available kernel descriptors. On the contrary, mAP of 71.18\% (on average) is achieved here as it is clear from Table 2 across all considered CNN configurations.

A comparative plot for mean average precision achieved by the state-of-the-art CNN architectures for the three databases (MINC-2500, FMD and ImageNet7) is shown in Figure 2. This plot encompasses only the six common/overlapping categories for the three databases to demonstrate the variations in performance of different CNN architectures across these categories. The Medium CNN architecture gives the best mAP (91.10\%) with stack augmentation both for training and for testing purposes. With the same configuration, the best performance of 77.46\% is achieved on ImageNet7, while a considerably lower mAP of 51.40\% is obtained for FMD. Such decline occurs primarily because of the limited dataset size, whereby learning the millions of parameters of a CNN is usually impractical and may lead to over-fitting.

\section{\uppercase{CROSS-DATASET ANALYSIS WITH REAL-WORLD IMAGES}}
\noindent Results for three different cross-dataset experiments are given in Table 3: 1) Training on FMD and testing on ImageNet7 2) Training on FMD and testing on MINC-2500 3) Training on MINC-2500 and testing on ImageNet7. Considering the fact that the FMD dataset is quite small, with only 100 images per material class, it performs better when used for training with reduced feature dimensionality per image, also observed in \cite{19}. In Table 3, with FMD as training database, the material classification pipeline performs best in testing the overlapping categories with ImageNet7 when Medium CNN architecture is used with 128 feature points per image extracted. The crop and flip augmentation and sum pooling collation is also used in this configuration and a mAP of ~82\% is achieved.
For FMD as training and MINC-2500 as testing database, the material classification pipeline achieves the best accuracy in testing the overlapping categories when CNN-M architecture is utilised with 2048 feature points per image extracted. Crop and flip augmentation and sum pooling are also used and the resulting mAP is ~76\%. It is evident from Table 3 that the performance of the system increases when MINC- 2500 is used as training database and overlapping categories of ImageNet7 are tested. This is due to the fact that MINC-2500 database enables the use of more images for positive training when testing the overlapping categories with ImageNet7. In this case, the highest accuracy is again achieved when CNN-M is used. However, only flip is used as augmentation and no collation is utilised with this CNN architecture as opposed to the above two cases. The resulting accuracy of the system is ~91\%. This is the case of finding the best balance before over-fitting occurs. Finally, the resulting average across all three experiments is ~82\%.

\begin{table*}[!t]
\centering
\caption{Cross-dataset material classification results. Training and testing are performed using 3 different databases of real-world images. The name on the top denotes the training database, while the name on the bottom implies the testing database. Bold font highlights the leading mean result for every experiment.}
\label{my-label}
\begin{tabular}{cccc|ccc}
\hline
\rowcolor[HTML]{C0C0C0} 
\cellcolor[HTML]{C0C0C0}                                  & \multicolumn{3}{c|}{\cellcolor[HTML]{C0C0C0}}                                      & \textbf{\begin{tabular}[c]{@{}c@{}}FMD\\ ImageNet7\end{tabular}} & \textbf{\begin{tabular}[c]{@{}c@{}}FMD\\ MINC-2500\end{tabular}} & \textbf{\begin{tabular}[c]{@{}c@{}}MINC-2500\\ ImageNet7\end{tabular}} \\
\rowcolor[HTML]{C0C0C0} 
\multirow{-2}{*}{\cellcolor[HTML]{C0C0C0}\textbf{Method}} & \multicolumn{3}{c|}{\multirow{-2}{*}{\cellcolor[HTML]{C0C0C0}\textbf{Image Aug.}}} & \textbf{mAP}                                                     & \textbf{mAP}                                                     & \textbf{mAP}                                                           \\
(a) CNN F                                                 & (C)                          & f                        & s                        & 78.23                                                            & 71.87                                                            & 85.11                                                                  \\
(b) CNN S                                                 & (C)                          & f                        & s                        & 83.50                                                            & 72.95                                                            & 86.18                                                                  \\ \hline
(c) CNN M                                                 & -                            &                          &                          & 82.40                                                            & 73.06                                                            & 87.64                                                                  \\
(d) CNN M                                                 & (C)                          & f                        & s                        & 81.68                                                            & 74.82                                                            & 85.79                                                                  \\
(e) CNN M                                                 & (C)                          & f                        & m                        & 81.69                                                            & 75.46                                                            & 86.55                                                                  \\
(f) CNN M                                                 & (C)                          & s                        & s                        & 79.52                                                            & 73.56                                                            & 89.88                                                                  \\
(g) CNN M                                                 & (C)                          & t                        & t                        & 80.22                                                            & 74.19                                                            & 89.53                                                                  \\
(h) CNN M                                                 & (C)                          & f                        & -                        & 80.31                                                            & 73.83                                                            & 82.71                                                                  \\
(i) CNN M                                                 & (F)                          & f                        & -                        & 81.91                                                            & 73.01                                                            & \textbf{91.03}                                                         \\
(j) CNN M GS                                              & -                            &                          &                          & 71.82                                                            & 66.78                                                            & 89.37                                                                  \\
(k) CNN M GS                                              & (C)                          & f                        & s                        & 75.95                                                            & 69.05                                                            & 87.87                                                                  \\ \hline
(l) CNN M 2048                                            & (C)                          & f                        & s                        & 80.27                                                            & \textbf{76.35}                                                   & 86.82                                                                  \\
(m) CNN M 1024                                            & (C)                          & f                        & s                        & 82.55                                                            & 74.85                                                            & 87.89                                                                  \\
(n) CNN M 128                                             & (C)                          & f                        & s                        & \textbf{82.90}                                                   & 73.99                                                            & 88.13                                                                 
\end{tabular}
\end{table*}

\section{\uppercase{CONCLUSIONS}}
\noindent We have performed a rigorous empirical evaluation of state-of-the-art CNN-based approaches for the material classification task. Out of the three baseline CNN architectures considered, it is evident that the Medium CNN architecture in general performs the best in combination with different data augmentation strategies for the three widely used material databases (FMD, MINC-2500 and ImageNet7). It will be an interesting future direction to investigate if synthetic data can be combined with real images to improve accuracy and generalisation abilities of CNNs \cite{25}.

\section*{AKNOWLEDGEMENTS}
\noindent We acknowledge MoD/Dstl and EPSRC for providing the grant to support the UK academics (Ales Leonardis) involvement in a Department of Defense funded MURI project. This work was also supported in part by EU H2020 RoMaNS 645582, EPSRC EPC EP/M026477/1 and ES/M010236/1.


\bibliographystyle{apalike}
{\small
\bibliography{example}}

\begin{thebibliography}{}

\bibitem[Bell et~al., 2015]{24}
Bell, S., Upchurch, P., Snavely, N., and Bala, K. (2015).
\newblock Material recognition in the wild with the materials in context
  database.
\newblock In {\em {IEEE} Conference on Computer Vision and Pattern Recognition,
  {CVPR}, 2015, Boston, MA, USA, June 7-12, 2015}, pages 3479--3487.

\bibitem[Chatfield et~al., 2014]{20}
Chatfield, K., Simonyan, K., Vedaldi, A., and Zisserman, A. (2014).
\newblock Return of the devil in the details: Delving deep into convolutional
  nets.
\newblock In {\em British Machine Vision Conference, {BMVC} 2014, Nottingham,
  UK, September 1-5, 2014}.

\bibitem[Cimpoi et~al., 2014]{16}
Cimpoi, M., Maji, S., Kokkinos, I., Mohamed, S., and Vedaldi, A. (2014).
\newblock Describing textures in the wild.
\newblock In {\em 2014 {IEEE} Conference on Computer Vision and Pattern
  Recognition, {CVPR} 2014, Columbus, OH, USA, June 23-28, 2014}, pages
  3606--3613.

\bibitem[Csurka et~al., 2004]{1}
Csurka, G., Bray, C., Dance, C., and Fan, L. (2004).
\newblock Visual categorization with bags of keypoints.
\newblock {\em Workshop on Statistical Learning in Computer Vision, ECCV},
  pages 1--22.

\bibitem[Dana et~al., 1999]{11}
Dana, K.~J., van Ginneken, B., Nayar, S.~K., and Koenderink, J.~J. (1999).
\newblock Reflectance and texture of real-world surfaces.
\newblock {\em ACM Trans. Graph.}, 18(1):1--34.

\bibitem[Deng et~al., 2009]{23}
Deng, J., Dong, W., Socher, R., Li, L., Li, K., and Li, F. (2009).
\newblock Imagenet: {A} large-scale hierarchical image database.
\newblock In {\em 2009 {IEEE} Computer Society Conference on Computer Vision
  and Pattern Recognition {(CVPR} 2009), 20-25 June 2009, Miami, Florida,
  {USA}}, pages 248--255.

\bibitem[Donahue et~al., 2014]{7}
Donahue, J., Jia, Y., Vinyals, O., Hoffman, J., Zhang, N., Tzeng, E., and
  Darrell, T. (2014).
\newblock Decaf: {A} deep convolutional activation feature for generic visual
  recognition.
\newblock In {\em Proceedings of the 31th International Conference on Machine
  Learning, {ICML} 2014, Beijing, China, 21-26 June 2014}, pages 647--655.

\bibitem[Fritz et~al., 2004]{17}
Fritz, M., Hayman, E., Caputo, B., and olof Eklundh, J. (2004).
\newblock {THE} {KTH}-{TIPS} database.

\bibitem[Girshick et~al., 2014]{29}
Girshick, R., Donahue, J., Darrell, T., and Malik, J. (2014).
\newblock Rich feature hierarchies for accurate object detection and semantic
  segmentation.
\newblock In {\em The IEEE Conference on Computer Vision and Pattern
  Recognition (CVPR)}.

\bibitem[Girshick et~al., 2013]{10}
Girshick, R.~B., Donahue, J., Darrell, T., and Malik, J. (2013).
\newblock Rich feature hierarchies for accurate object detection and semantic
  segmentation.
\newblock {\em CoRR}, abs/1311.2524.

\bibitem[Hu et~al., 2011]{15}
Hu, D., Bo, L., and Ren, X. (2011).
\newblock Toward robust material recognition for everyday objects.
\newblock In {\em British Machine Vision Conference, {BMVC} 2011, Dundee, UK,
  August 29 - September 2, 2011. Proceedings}, pages 1--11.

\bibitem[Huang et~al., 2011]{27}
Huang, Y., Huang, K., Yu, Y., and Tan, T. (2011).
\newblock Salient coding for image classification.
\newblock In {\em Computer Vision and Pattern Recognition (CVPR), 2011 IEEE
  Conference on}, pages 1753--1760. IEEE.

\bibitem[Jia et~al., 2014]{22}
Jia, Y., Shelhamer, E., Donahue, J., Karayev, S., Long, J., Girshick, R.,
  Guadarrama, S., and Darrell, T. (2014).
\newblock Caffe: Convolutional architecture for fast feature embedding.
\newblock In {\em Proceedings of the 22Nd ACM International Conference on
  Multimedia}, MM '14, pages 675--678, New York, NY, USA. ACM.

\bibitem[Krizhevsky et~al., 2012]{3}
Krizhevsky, A., Sutskever, I., and Hinton, G.~E. (2012).
\newblock Imagenet classification with deep convolutional neural networks.
\newblock In {\em Advances in Neural Information Processing Systems 25: 26th
  Annual Conference on Neural Information Processing Systems 2012. Proceedings
  of a meeting held December 3-6, 2012, Lake Tahoe, Nevada, United States.},
  pages 1106--1114.

\bibitem[LeCun et~al., 1989]{2}
LeCun, Y., Boser, B., Denker, J.~S., Henderson, D., Howard, R.~E., Hubbard, W.,
  and Jackel, L.~D. (1989).
\newblock Backpropagation applied to handwritten zip code recognition.
\newblock {\em Neural Computation}, 1(4):541--551.

\bibitem[Liu et~al., 2010]{14}
Liu, C., Sharan, L., Adelson, E.~H., and Rosenholtz, R. (2010).
\newblock Exploring features in a bayesian framework for material recognition.
\newblock In {\em The Twenty-Third {IEEE} Conference on Computer Vision and
  Pattern Recognition, {CVPR} 2010, San Francisco, CA, USA, 13-18 June 2010},
  pages 239--246.

\bibitem[Oquab et~al., 2014]{8}
Oquab, M., Bottou, L., Laptev, I., and Sivic, J. (2014).
\newblock Learning and transferring mid-level image representations using
  convolutional neural networks.
\newblock In {\em Proceedings of the 2014 IEEE Conference on Computer Vision
  and Pattern Recognition}, CVPR '14, pages 1717--1724, Washington, DC, USA.
  IEEE Computer Society.

\bibitem[Razavian et~al., 2014]{9}
Razavian, A.~S., Azizpour, H., Sullivan, J., and Carlsson, S. (2014).
\newblock {CNN} features off-the-shelf: An astounding baseline for recognition.
\newblock In {\em Proceedings of the 2014 IEEE Conference on Computer Vision
  and Pattern Recognition Workshops}, CVPRW '14, pages 512--519, Washington,
  DC, USA. IEEE Computer Society.

\bibitem[Sermanet et~al., 2013]{5}
Sermanet, P., Eigen, D., Zhang, X., Mathieu, M., Fergus, R., and LeCun, Y.
  (2013).
\newblock Overfeat: Integrated recognition, localization and detection using
  convolutional networks.
\newblock {\em CoRR}, abs/1312.6229.

\bibitem[Sharan et~al., 2010]{13}
Sharan, L., Rosenholtz, R., and Adelson, E. (2010).
\newblock Material perception: What can you see in a brief glance?
\newblock {\em Journal of Vision}, 9(8):784--784a.

\bibitem[Simonyan and Zisserman, 2014]{6}
Simonyan, K. and Zisserman, A. (2014).
\newblock Two-stream convolutional networks for action recognition in videos.
\newblock In {\em Advances in Neural Information Processing Systems 27: Annual
  Conference on Neural Information Processing Systems 2014, December 8-13 2014,
  Montreal, Quebec, Canada}, pages 568--576.

\bibitem[Varma and Zisserman, 2009]{12}
Varma, M. and Zisserman, A. (2009).
\newblock A statistical approach to material classification using image patch
  exemplars.
\newblock {\em IEEE Transactions on Pattern Analysis and Machine Intelligence},
  31(11):2032--2047.

\bibitem[Vedaldi and Zisserman, ]{21}
Vedaldi, A. and Zisserman, A.
\newblock Recognition of object categories practical.

\bibitem[Wang et~al., 2010]{28}
Wang, J., Yang, J., Yu, K., Lv, F., Huang, T., and Gong, Y. (2010).
\newblock Locality-constrained linear coding for image classification.
\newblock In {\em Computer Vision and Pattern Recognition (CVPR), 2010 IEEE
  Conference on}, pages 3360--3367.

\bibitem[Weinmann et~al., 2014]{25}
Weinmann, M., Gall, J., and Klein, R. (2014).
\newblock Material classification based on training data synthesized using a
  {BTF} database.
\newblock In {\em Computer Vision - {ECCV} 2014 - 13th European Conference,
  Zurich, Switzerland, September 6-12, 2014, Proceedings, Part {III}}, pages
  156--171.

\bibitem[Zeiler and Fergus, 2014]{4}
Zeiler, M.~D. and Fergus, R. (2014).
\newblock Visualizing and understanding convolutional networks.
\newblock In {\em Computer Vision - {ECCV} 2014 - 13th European Conference,
  Zurich, Switzerland, September 6-12, 2014, Proceedings, Part {I}}, pages
  818--833.

\bibitem[Zheng et~al., 2014]{19}
Zheng, S., Cheng, M., Warrell, J., Sturgess, P., Vineet, V., Rother, C., and
  Torr, P. H.~S. (2014).
\newblock Dense semantic image segmentation with objects and attributes.
\newblock In {\em 2014 {IEEE} Conference on Computer Vision and Pattern
  Recognition, {CVPR} 2014, Columbus, OH, USA, June 23-28, 2014}, pages
  3214--3221.

\end{thebibliography}

\end{document}